\documentclass{Interspeech2024}

\usepackage{acronym}
\newacro{ASR}[ASR]{Automatic Speech Recognition}
\newacro{WER}[WER]{Word Error Rate}
\newacro{CER}[CER]{Character Error Rate}
\newacro{MER}[MER]{Match Error Rate}
\newacro{WIL}[WIL]{Word Information Lost}
\newacro{WWER}[WWER]{Weighted Word Error Rate}
\newacro{ACE}[ACE]{Automated-Caption Evaluation}
\newacro{AI}[AI]{artificial intelligence}
\newacro{FCC}[FCC]{Federal Communications Commission}
\newacro{E2E}[E2E]{end-to-end}
\newacro{DNN}[DNN]{deep neural network}
\newacro{SER}[SER]{Slot Error Rate}
\newacro{NLP}[NLP]{natural language processing}
\newacro{NER}[NER]{Named-entity recognition}
\newacro{POS}[POS]{part-of-speech}

\newcommand\scalemath[2]{\scalebox{#1}{\mbox{\ensuremath{\displaystyle #2}}}}

\interspeechcameraready

\title{Beyond Levenshtein: Leveraging Multiple Algorithms for Robust Word Error Rate Computations And Granular Error Classifications}

\name[affiliation={1,2}]{Korbinian}{Kuhn}
\name[affiliation={1}]{Verena}{Kersken}
\name[affiliation={1}]{Gottfried}{Zimmermann}

\address{
  $^1$ Stuttgart Media University, Germany\\
  $^2$ University of Tübingen, Germany
}
\email{kuhnko@hdm-stuttgart.de, kersken@hdm-stuttgart.de, gzimmermann@acm.org}


\begin{document}

\maketitle

\begin{abstract}
The Word Error Rate (WER) is the common measure of accuracy for Automatic Speech Recognition (ASR). Transcripts are usually pre-processed by substituting specific characters to account for non-semantic differences. As a result of this normalisation, information on the accuracy of punctuation or capitalisation is lost. We present a non-destructive, token-based approach using an extended Levenshtein distance algorithm to compute a robust WER and additional orthographic metrics. Transcription errors are also classified more granularly by existing string similarity and phonetic algorithms. An evaluation on several datasets demonstrates the practical equivalence of our approach compared to common WER computations. We also provide an exemplary analysis of derived use cases, such as a punctuation error rate, and a web application for interactive use and visualisation of our implementation. The code is available open-source.\footnote{https://github.com/shuffle-project/beyond-levenshtein}
\end{abstract}

\begin{figure*}[ht!]
  \centering
  \includegraphics[width=\linewidth]{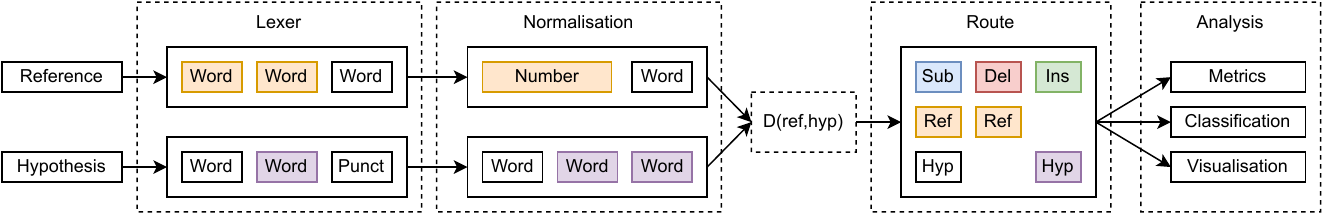}
  \caption{\textbf{Processing pipeline}: The lexer transforms the input texts into a list of tokens, which are further normalised by several text pre-processors. An extended variant of the Levenshtein distance algorithm with compound word detection and variable edit costs determines the shortest route of modifications. Substitutions are further classified as punctuation, capitalisation, or word errors (e.g. suffix or homophone). The route is used for calculating metrics like WER, categorising types of errors, and visualising text differences.}
  \label{fig:processing}
\end{figure*}

\section{Introduction}

Transcription accuracy is a central performance evaluation of \ac{ASR} models, because it directly measures their ability to convert spoken language into accurate text representations. However, the complexity of speech and language poses challenges in determining accuracy and evaluations extend beyond merely counting errors. Evaluating transcription accuracy involves rating the severity of errors, measuring the impact of punctuation or capitalisation discrepancies, and assessing how errors influence the human-perceived understandability of a text. A critical aspect in the evaluation process is the need for automated metrics that can efficiently assess large datasets to demonstrate the robustness of an \ac{ASR} model under diverse linguistic scenarios and environmental conditions \cite{Geirhos2020}. Metrics may also be tailored towards specific use cases, such as the correct transcription of numbers and addresses \cite{Aksenova2021}.

The standard metric to report accuracy in \ac{ASR} research is the \ac{WER} \cite{Baevski2020, Zhang2022, Radford2023}. It represents the average number of transcription errors per 100 words. The underlying algorithm is the Levenshtein distance, which calculates the minimum number of operations (insertions, deletions, or substitutions) needed to transform one string into another \cite{Levenshtein1966}. The \ac{WER} is computed by determining the minimum edit distance on a word level to measure the number of modifications between an \ac{ASR} generated hypothesis transcript and a manually created "error-free" reference. Alternatively, for languages that do not use spaces between words (e.g. Chinese), the \ac{CER} is used.

To make the \ac{WER} more reliable, it is necessary to reduce errors resulting from non-semantic text differences like capitalisation, abbreviations, or numerical notations. Transcripts are commonly pre-processed before the \ac{WER} is calculated. Standard normalisation removes newline characters, repeated spaces, punctuation, and capitalisation. Unifying abbreviations, common contractions, spelling differences, and written numbers can further reduce non-semantic differences \cite{Koenecke2020, Radford2023}. A drawback of these modifications is that they eliminate criteria affecting the readability of a transcript. Reading \ac{ASR}-generated text can be challenging and is further complicated by the presence of orthographic errors \cite{Butler2019}. While the \ac{FCC}\footnote{Federal Communications Commission. (2014, Feb. 24). Closed Captioning Quality Report and Order, Declaratory Ruling, FNPRM.} mandates correct punctuation and capitalisation, the \ac{WER} is limited in its ability to measure their accuracy.

Especially in the field of accessibility, the \ac{WER} has been criticised as a valid measure of accuracy because the importance of individual words to the overall understandability of a text is not considered, and all errors are penalised equally. Several studies report a weak correlation between \ac{WER} and understandability ratings by human readers \cite{Wang2003, Mishra2011, Favre2013}. Other measures, such as \ac{MER} and \ac{WIL} \cite{Morris2004} also rely on the Levenshtein distance, and have similar problems as the \ac{WER}. General machine translation metrics like BLEU\cite{Papineni2002} or METEOR\cite{Banerjee2005} are less specific for the evaluation of transcription accuracy, but are often reported complementary to the \ac{WER}, e.g. to reflect the multilingual capabilities of an \ac{ASR} model. Although there is no established automated method for calculating punctuation or capitalisation errors, the precision of specific aspects can be quantified using F1-scores or the \ac{SER} \cite{Pais2022}.

More recent approaches use \ac{AI} to classify the importance of words and the severity of errors \cite{Amin2023}. However, a machine learning based metric can inherit biases present in its training data, be susceptible to overfitting, or adapt poorly to domain shifts. \ac{ACE}\cite{Kafle2017} and its successor ACE2\cite{Kafle2019}, for example, are trained on conversational speech and are less suitable for the evaluation of live television captioning \cite{Wells2022}. Metrics such as the NER-model\cite{Romero2015} and the \ac{WWER}\footnote{T. Apone, M. Brooks, T. O’Connell, "Caption Accuracy Metrics Project," 2010.}, require manual effort to rate the severity of each transcription error. Although manual evaluations can be helpful for specific use cases, they lack objectivity, cannot be replicated, and are hard to obtain for large datasets.

While there is an ongoing search for alternative metrics, the \ac{WER} remains the commonly used one. We propose an extended Levenshtein distance algorithm that allows the computation of a robust \ac{WER} while preserving punctuation and capitalisation. We further utilise commonly used algorithms in the field of \ac{NLP} to classify transcription errors more granularly. We see the following applications of our approach for future research:

\begin{itemize}
    \item Calculation of individual or combined metrics based on error types (e.g. word, punctuation, capitalisation, number, ...).
    \item Detection of performance increasements and negative side effects of \ac{ASR} models complementary to the overall \ac{WER}.
    \item Visual analysis of transcription errors and text normalisations through a graphical web interface.
\end{itemize}

\section{Error Rate Computation}

The process is visualised in Fig. \ref{fig:processing} and explained in detail in the following section.

\subsection{Lexer}

In typical \ac{WER} computations, the input strings are split into parts at each space or newline character, e.g. with a regular expression. In contrast, a lexer can decide on more complex rules based on the preceding and following characters. We use a lexer that parses a string into categorised tokens such as words, numbers, punctuation or symbols. Each token contains the raw value with all characters and a normalised value used to compare the equality of two tokens. Non-word characters, such as spaces, tabs, newline characters, em-dashes, or preceding quotation marks are stored as a prefix or suffix of a token. A new token is created if a word character appears after a suffix. Punctuation characters also create new tokens, except for a few exceptions, e.g. when a period is at the end of an abbreviation (like Mrs.) or between two digits (like 3.14). Some common symbols, like per cent or currency signs, also create a new token.

\subsection{Normalisation}

The initial list of tokens is then processed through various normalisers, which can modify, split, or merge tokens. For example, if a common contraction like "won’t" is detected, a token is split into two word tokens "will" and "not". If a token’s value is changed, the name of the normalisation is added to that token’s list of normalisations for a later analysis. All normalisations are non-destructive, and the original (unmodified) value is stored with the token. The normalisers further extend existing solutions of JiWER\footnote{https://github.com/jitsi/jiwer} and Whisper\footnote{https://github.com/openai/whisper}.

The following normalisations are applied: Common abbreviations (e.g. "Mr." or "etc.") and contractions (e.g. "I’m" or "gonna") are replaced with their long form. Annotations within parenthesis (e.g. \textless unknown\textgreater, (pause), [unintelligible]) and interjections (e.g. "hmm", "um") are removed. Spelling differences between UK- and US-English are unified to the US variant (e.g. "analyse", "colour"). Diacritics are replaced with ASCII letters (e.g. "ä"). Symbols for currencies and per cent are replaced with their literal expression. The notation of numbers and currencies can lead to multiple non-semantic errors without normalisation (e.g. "two thousand dollars" and "\$2000"). As the normalisation from text to numbers is not trivial, we reimplemented the number normaliser from Whisper, even though it is not error-free.

\subsection{Extended Levenshtein Distance}

We extend the Levenshtein distance to detect compound words and use variable costs per operation depending on the involved token types. Algorithm is defined as follows:

% d(a,b)
\begin{equation}
\scalemath{0.8}{
  d_{a,b}(i,j) = min
    \begin{cases}
        0 & \text{if $i=j=0$}\\
        d_{a,b}(i-1,j-1) & \text{if $a_i = b_j$} \\
        d_{a,b}(i-x,j-y) & \text{if $a_{[i-x:i]} = b_{[j-y:j]}$}  \\
        d_{a,b}(i-1,j) + cost(a_i) & \text{if $i>0$}\\
        d_{a,b}(i,j-1) + cost(b_j) & \text{if $j>0$}\\
        d_{a,b}(i-1,j-1) + cost(a_i,b_j) \\
    \end{cases}       
}
\label{eq:levenshtein}
\end{equation}

where $d_{a,b}$ is the recursively defined distance function of two token lists $a$ and $b$. $i$ and $j$ index into the respective list. $a_{[i-x:i]}$ represents a slice of $a$ with length $x$ starting at index $i-x$, to capture compound words. The condition $1 \leq x \leq min(i, K)$ must hold, where $K$ is the maximum number of compounds. $cost$ is defined depending on the number of parameters as:

% costs(c)
\begin{equation}
\scalemath{0.8}{
  cost(c_k) = 
    \begin{cases}
        0.5 & \text{if $c_k \in \mathbb{P}$}\\
        1  & \text{otherwise} \\
    \end{cases}   
}
\label{eq:cost-t}
\end{equation}
%

% costs(a,b)
\begin{equation}
\scalemath{0.8}{
  cost(a_i,b_j) = 
    \begin{cases}
        2 & \text{if $a_i \in \mathbb{A} \wedge b_j \in \mathbb{B}$ \text{where} $\mathbb{A} \cap \mathbb{B} = \varnothing $}\\
        0.5 & \text{if $a_i \in \mathbb{P} \wedge b_j \in \mathbb{P}$}\\
        0.5 & \text{if $\lfloor{a_{i}}\rfloor = \lfloor{b_{j}}\rfloor$}\\
        1 & \text{otherwise} \\
    \end{cases}  
}
\label{eq:cost-ab}
\end{equation}

where $cost(\cdot)$ is a function that returns the costs of insertion or deletion for one token and $cost(\cdot,\cdot)$ returns the costs for the substitution of two tokens depending on the token type. $\mathbb{A}$ and $\mathbb{B}$ are sets that contain all tokens of a type,  where $\mathbb{P}$ is the concrete set of all punctuation. We define $\mathbb{P} = \{.,!?;:\}$. $\lfloor c_k \rfloor$ converts all characters of $c_k$ to lowercase.

$K$ can be adjusted to improve computation speed in balance to the morpheme-to-word ratio of the applied language (lower for analytic and higher for synthetic languages). We used $K = \infty$ in our implementation.

\begin{figure}[ht]
  \centering
  \includegraphics[width=0.8\linewidth]{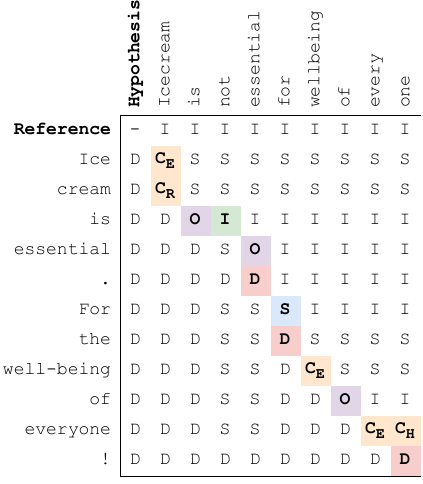}
  \caption{Backtrace Matrix: The shortest route is found by traversing the operations starting from the bottom right of the matrix. The operations and movements are: $O$=OK (up left), $I$=Insertion (left), $D$=Deletion (up), $S$=Substitution (up left), $C_H$=Compound hypothesis (left), $C_R$=Compound reference (up), $C_E$=Compound end (up left).}
  \label{fig:backtrace-matrix}
\end{figure}

The Levenshtein distance is calculated by iteratively filling a 2D matrix, which stores the optimal sequence of edits to transform one string into another (or, in our case, two lists of tokens). Each cell in the matrix corresponds to a specific position of the reference and the hypothesis. The matrix is of size ($len(ref) * len(hyp)$, where each cell ($i$, $j$) represents the cost of transforming the first $i$ tokens of the reference into the first $j$ tokens of the hypothesis. The value in each cell is the minimum cost among the possible operations.

Compound words are detected if one or more consecutive tokens of the hypothesis match one or more consecutive tokens of the reference (ignoring spaces and hyphens), for example, "icecream" and "ice-cream". Compounds will not be recognised if they are incorrectly separated by punctuation characters. In some cases this approach can also lead to false positives, for example, "a long" and "along". 

In contrast to the original Levenshtein distance, modification costs depend on the involved token types. Insertions, deletions, and substitutions of words are also penalised with 1, but capitalisation and punctuation errors are considered less critical and are weighted with 0.5. Substitutions between a word and a punctuation token are penalised with 2. Thus, removing a comma and inserting a word is preferred over substituting a comma to a word.

\subsection{Route}

After the matrix is filled, the best route (shortest path) is traversed by backtracing the operations matrix from the bottom right and following each operation (see Fig. \ref{fig:backtrace-matrix}). The determined route consists of elements with either a reference token (deletion), a hypothesis token (insertion), or both (ok, substitution, compound). Every substitution is further classified as punctuation or word by comparing the hypothesis and reference token values. Word substitutions are further classified as capitalisation, compound word, and number errors if possible. If the hypothesis value is partly included in the reference word, it is categorised as prefix, suffix, affix or with the Porter-Stemmer\cite{Porter1980} algorithm. Homophones are detected using the Double-Metaphone\cite{Philips2000} algorithm.

\subsection{Analysis}

\subsubsection{Metrics}

Punctuation and capitalisation accuracy is reported using \ac{SER} and F1-scores, as there is no standard metric for these evaluations \cite{Pais2022}.  \ac{WER} and \ac{SER} are defined as follows:

% WER/SER
\begin{equation}
  WER = SER = \frac{S + D + I}{C + S + D} 
\label{eq:wer}
\end{equation}

where $C$ is the number of correct tokens, $S$ the number of substitutions (mismatching tokens), $D$ the number of deletions (missing tokens), $I$ the number of insertions (incorrect predictions). 

Using the same parameters, F1-score is defined as follows:

% F1
\begin{equation}
  F1 = \frac{2C}{2C + 2S + D + I} 
\label{eq:f1}
\end{equation}

\subsubsection{Visualisation}

\begin{figure}[ht]
  \centering
  \frame{
  \includegraphics[width=\linewidth]{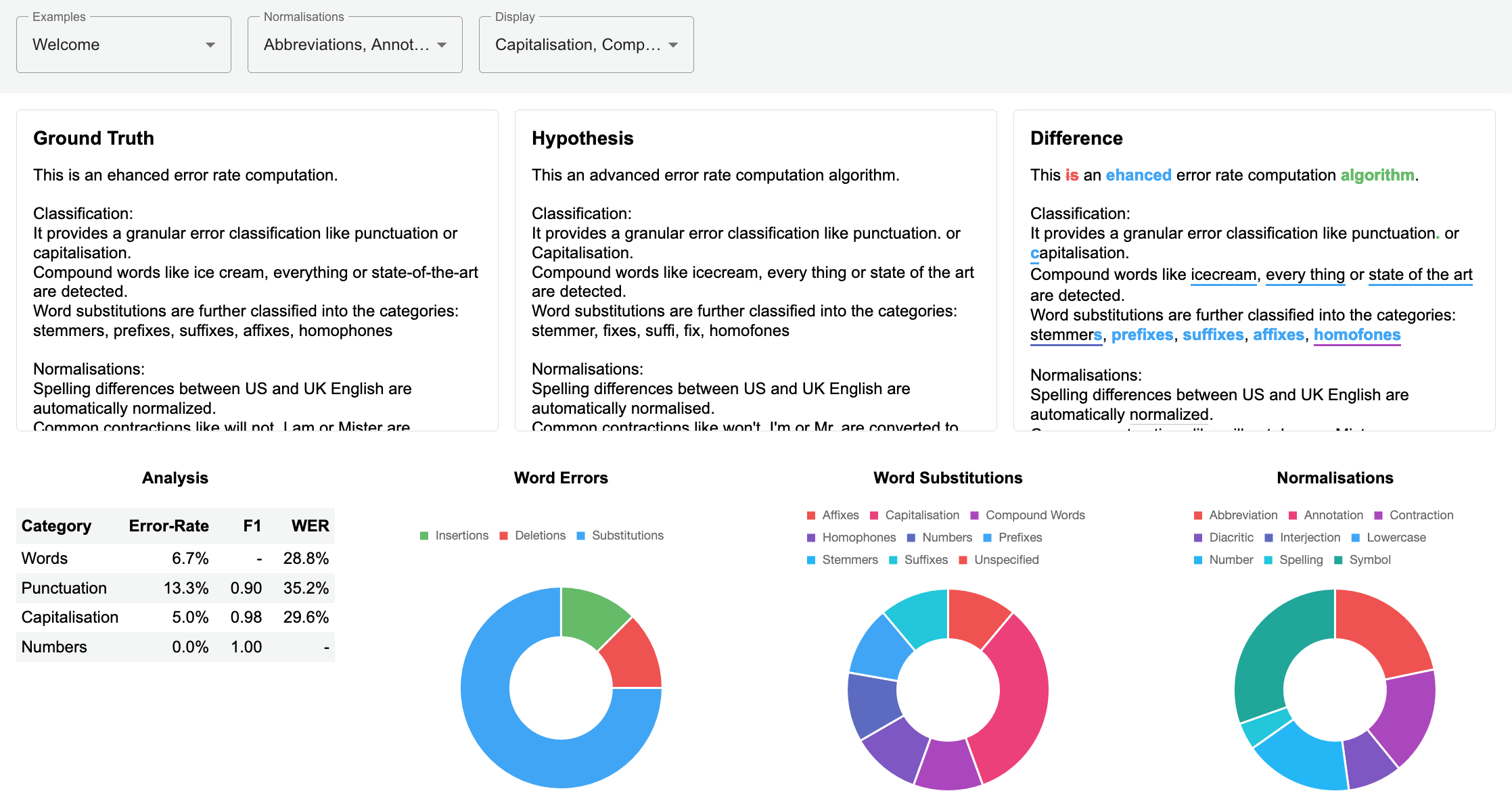}
  }
  \caption{An interactive web application visualises text differences, error types, and normalisations and calculates several error metrics like \ac{WER}, \ac{SER}, and F1-scores.}
  \label{fig:visualisation}
\end{figure}

An interactive web application is provided for testing the algorithm with pre-transcribed samples of multiple speech datasets or custom user input.\footnote{https://shuffle-project.github.io/beyond-levenshtein} The interface, shown in Fig. \ref{fig:visualisation}, visualises text differences, types of errors, and normalisations. Metrics such as the \ac{WER}, \ac{SER}, and F1-scores are displayed in comparison to the "traditional" \ac{WER}. Multiple graphs accumulate the classified substitution errors and the number of normalisations. All normalisations can be toggled to examine their impact on the calculated metrics.

\section{Evaluation}

We used eight open datasets for long-form English transcriptions covering different scenarios like book readings, colloquial speech, business meetings, and presentations: CORAAL\cite{Gunter2021}, Earnings-21\cite{DelRio2021}, Earnings-22\cite{DelRio2022}, Kincaid46\footnote{J. Kincaid, "Which Automatic Transcription Service is the Most Accurate?," 2018.}, LibriSpeech-PC\cite{Meister2023}, Meanwhile\cite{Radford2023}, Rev16\cite{Radford2023}, TED-LIUM 3\cite{Hernandez2018}. In total, these datasets contain 745 files and 349 hours of audio. 

Two popular \ac{ASR} frameworks were used to transcribe the datasets:  Whisper as an \ac{E2E} model (all English and multilingual models) and Vosk as a hybrid model using Kaldi (small-en-us-0.15, en-us-0.22). Punctuation and capitalisation of the Vosk transcripts were created with the vosk-recasepunc-en-0.22 model. In total, 11175 transcripts were generated.

\subsection{Word Error Rate Robustness}

We evaluated the extended Levenshtein distance algorithm by comparing three different \ac{WER} computations for all transcriptions: Ours (implemented as described in the previous section), Whisper (using the Whisper text normaliser and JiWER \ac{WER} computation), JiWER (using the JiWER text normaliser and \ac{WER} computation). 

Table \ref{tab:wer} shows the average \ac{WER} of all transcriptions per dataset. Our implementation reports the lowest \ac{WER} on average and on the most datasets. While the results are close to Whisper’s text normalisation, they are slightly higher for JiWER, which applies less extensive text pre-processing. The average standard deviation of ours ($SD=17.5$), Whisper ($SD=17.3$), and JiWER ($SD=17.3$) is similar.

\begin{table}[ht]
  \caption{Average Word Error Rate of all transcriptions for different computation libraries per dataset.}
  \label{tab:wer}
  \centering
  \begin{tabular}{lrrrr}
    \toprule
    Dataset & Ours & Whisper & JiWER  \\
    \midrule
    CORAAL & \textbf{38.2} & 38.4 & 40.2 \\
    Earnings-21 & \textbf{20.4} & 20.5 & 23.8 \\
    Earnings-22 & \textbf{27.6} & 27.7 & 31.4 \\
    Kincaid46 & 19.9 & \textbf{19.7} & 21.3 \\
    LibriSpeech-PC & \textbf{13.7} & 14.1 & 14.7 \\
    Meanwhile & \textbf{14.3} & 14.9 & 17.1 \\
    Rev16 & \textbf{19.1} & 19.2 & 20.4 \\
    TED-LIUM 3 & \textbf{8.9} & 9.0 & 10.2 \\
    \midrule
    Average & \textbf{25.7} & 26.0 & 27.9 \\
    \bottomrule
  \end{tabular}
\end{table}

A one-way ANOVA was conducted to determine the effect of the three different implementations on the resulting \ac{WER}. There was a significant effect in \ac{WER} between at least two implementations $[F(2, 33522) = 52.001, p < .001]$. A Tukey’s HSD test showed a significant difference between ours and JiWER ($p < .001$), and between Whisper and JiWER ($p < .001$). There was no significant difference between ours and Whisper ($p = .601$).

Bootstrap resampling with 10000 iterations was conducted to determine 95\% confidence intervals for the mean \ac{WER} difference between the three implementations \cite{Efron1994}. The mean difference between ours and Whisper was $-0.002$, $SD=0.007$, 95\% CI $[-0.002, -0.002]$, between ours and JiWER was $-0.022$, $SD=0.018$, 95\% CI $[-0.022,-0.021]$, and between Whisper and JiWER was $-0.019$, $SD=0.018$, 95\% CI $[-0.020,-0.019]$.

\subsection{Punctuation And Capitalisation}

As an exemplary use case, table \ref{tab:wer-ser-f1} shows the average \ac{WER} of different \ac{ASR} models and complementary metrics for punctuation and capitalisation accuracy. Although the \ac{WER} of the hybrid \ac{ASR} model vosk is slightly lower than Whisper’s base.en model, the post-processed punctuation and capitalisation are less accurate. A comparison of Whisper’s largest English model with the multilingual large-v3 model shows that the \ac{WER} is lower for the multilingual model, but the English model performs better on the other metrics.

\setlength{\tabcolsep}{4pt}
\begin{table}[ht]
  \caption{Metrics for word, punctuation, and capitalisation errors of different \ac{ASR}-models.}
  \label{tab:wer-ser-f1}
  \centering
  \begin{tabular}{lccccc}
    \toprule
     & Words & \multicolumn{2}{c}{Punctuation} & \multicolumn{2}{c}{Capitalisation}  \\
     Model & WER$\downarrow$ & SER$\downarrow$ & F1$\uparrow$ & SER$\downarrow$ & F1$\uparrow$ \\
    \midrule
    vosk-en-us-0.22 & \textbf{24.6} & 80.5 & 0.51 & 59.2 & 0.72 \\
    whisper-base.en & 25.1 & \textbf{48.8} & \textbf{0.66} & \textbf{35.8} & \textbf{0.82} \\
    \midrule
    whisper-medium.en & 19.4 & \textbf{43.4} & \textbf{0.68} & \textbf{32.3} & \textbf{0.83} \\
    whisper-large-v3 & \textbf{15.9} & 48.5 & 0.61 & 60.6 & 0.70 \\
    \bottomrule
  \end{tabular}
\end{table}
\setlength{\tabcolsep}{6pt}

\section{Conclusion}

Despite the ongoing search for alternative measures, \ac{WER} is the standard metric for reporting the accuracy of \ac{ASR}. Current \ac{ASR}-models can achieve very low error rates, which increases the interest in evaluating additional criteria such as punctuation or capitalisation accuracy. However, the \ac{WER} is limited in its ability to measure their accuracy. We presented an extended Levenshtein distance algorithm to handle punctuation, capitalisation, and compound words. The token-based approach preserves the original text and applies text pre-processing non-destructively.

A statistical analysis using several long-form English datasets showed that our implementation computes similar \ac{WER} values as commonly used libraries. The results indicate that the calculated \ac{WER} does not differ significantly between the edit-distance algorithms, but rather between the amount of text pre-processing. Given the small difference between ours and Whisper’s, it is reasonable to conclude that the two implementations demonstrate practical equivalence in computing \ac{WER}. Thus, the presented algorithm can be used alternatively to calculate a robust \ac{WER}, whilst allowing the computation of additional metrics such as punctuation accuracy.

An exemplary analysis of these metrics showed that two \ac{ASR}-models can achieve a similar \ac{WER} but differ significantly in punctuation and capitalisation accuracy. Further research is required to test the practicability of these additional measures for evaluating and optimising \ac{ASR}-models and if they can support the definition of accessibility requirements for \ac{ASR}-generated transcriptions.

\section{Acknowledgements}

We thank Daniel Grießhaber for his valuable thoughts on the extended Levenshtein distance algorithm. This work was conducted as part of the SHUFFLE Project and funded by “Stiftung Innovation in der Hochschullehre”.

\bibliographystyle{IEEEtran}
\bibliography{mybib}

\end{document}